\documentclass[11pt]{amsart}

\usepackage[a4paper,margin=1in]{geometry}
\usepackage{amsmath,amssymb,amsthm,mathtools}
\usepackage{booktabs}
\usepackage{array}
\usepackage{enumitem}
\usepackage{xcolor}
\usepackage{float}
\usepackage{graphicx}
\usepackage{listings}
\usepackage{xurl}
\usepackage[hidelinks]{hyperref}
\lstdefinestyle{lean}{
  basicstyle=\ttfamily\small,
  breaklines=true,
  columns=fullflexible,
  keepspaces=true
}

\lstdefinestyle{shell}{
  basicstyle=\ttfamily\small,
  breaklines=true,
  columns=fullflexible,
  keepspaces=true
}

\newtheorem{theorem}{Theorem}

\theoremstyle{definition}

\newtheorem{remark}[theorem]{Remark}


\title[AI-ASSISTED THEOREM PROVING]{Using the Aristotle API for AI-Assisted Theorem Proving in Lean 4: A Formalisation Case Study of the Grasshopper Problem}

\author[Gabriel R. Lau]{Gabriel R. Lau}

\makeatletter
\def\@setauthors{%
  \begingroup
  \def\thanks{\protect\thanks@warning}%
  \trivlist
  \centering\footnotesize \@topsep30\p@\relax
  \advance\@topsep by -\baselineskip
  \item\relax
  \author@andify\authors
  \def\\{\protect\linebreak}%
  \MakeUppercase{\authors}\par
  \vspace{0.35em}%
  {\normalfont\small Nanyang Technological University, Singapore}\par
  {\normalfont\small\texttt{gabriel.laury@ntu.edu.sg}}%
  \ifx\@empty\contribs
  \else
    ,\penalty-3 \space \@setcontribs
    \@closetoccontribs
  \fi
  \endtrivlist
  \endgroup
}
\makeatother

\date{May 2026}

\begin{document}

\begin{abstract}
AI-assisted theorem proving can now generate substantial Lean developments for olympiad-level mathematics, but the evidential status of such developments depends on which declarations are actually verified. This paper reports a Lean 4 formalisation case study of an Aristotle API proof attempt for the Grasshopper problem, originally posed as IMO 2009 Problem 6. The generated artefact states a generalised Lean version of the theorem, contains four verified helper lemmas for local components of a maximality and adjacent-swap exchange strategy, and leaves the main theorem \texttt{grasshopper} closed directly by one unresolved \texttt{sorry}. The verified components establish that the final partial sum equals the total sum, that an adjacent transposition can affect only the relevant intermediate partial sum, that the changed partial sum has the expected form, and that maximality at a position admitting an adjacent successor swap forces a corresponding forbidden-set membership fact. The Aristotle output summary identifies the intended remaining mathematical step as the global counting step needed to show that these membership facts produce at least \(n\) distinct forbidden values, contradicting the cardinality assumption \(|M| < n\); the Lean source itself does not reduce the main theorem to a separately encoded counting lemma. This case study gives an inspectable example of a central limitation in AI-assisted formalisation, namely that local proof search can succeed while the global combinatorial bookkeeping required for a theorem remains unresolved. The paper contributes a reproducible Lean artefact and a precise analysis of its verified and unverified proof content.
\end{abstract}

\maketitle

\section{Introduction}

AI-assisted theorem proving raises a practical problem of proof interpretation. When an AI system generates a Lean development, one must distinguish between declarations that are fully verified and declarations that are accepted only because they contain unresolved proof placeholders. Since Lean accepts declarations closed by \texttt{sorry}, successful compilation alone is insufficient evidence of a completed proof. In Lean, \texttt{sorry} is a placeholder that allows a declaration to be accepted without a completed proof. A theorem closed by \texttt{sorry} should therefore not be treated as a machine-checked proof~\cite{leanrefsorry}. This paper examines that distinction through a Lean 4 formalisation attempt for the Grasshopper problem, originally posed as IMO 2009 Problem 6.

The Grasshopper problem was posed as Problem 6 of the 50th International Mathematical Olympiad, held in Bremen, Germany, in July 2009. The official IMO individual-results table records contestant-level scores for all six problems. Aggregating the Problem 6 scores shows that it was the least solved problem of the competition, with mean score 0.168 out of 7 and 540 contestants receiving zero points~\cite{imo2009stats}. Contemporary discussion by Tao treated the problem as a suitable subject for collaborative exploration~\cite{tao2009minipolymath}. Related work of K\'{o}s studies a signed-jump variant and contrasts the original problem with variants for which polynomial-method arguments become available~\cite{kos2011signed}.

This case study is situated within recent work combining language-model-guided search with formal proof assistants. The miniF2F benchmark was introduced to evaluate formal olympiad-level mathematics across proof-assistant systems~\cite{zheng2022minif2f}. DeepSeek-Prover uses large-scale synthetic Lean data to improve theorem proving by language models~\cite{xin2024deepseekprover}. AlphaGeometry combines neural guidance with symbolic deduction for olympiad geometry problems~\cite{trinh2024alphageometry}. Aristotle is described in its technical report as an automated theorem-proving system using Lean proof search, informal lemma generation and formalisation, reinforcement-learning improvements, and a separate geometry solver~\cite{harmonic2025aristotle}. The present paper studies a narrower artefact, a partial Aristotle-generated Lean 4 proof development for one combinatorial problem.

The paper makes two contributions. First, it reports a case study of AI-assisted theorem proving in Lean 4 using the Aristotle API on a difficult olympiad combinatorics problem. Second, it illustrates a current limitation of AI-assisted formalisation, namely the gap between locally verified proof components and the broader mathematical reasoning needed to complete a proof. The case study interprets the generated Lean development by its verified declarations, dependencies, and unresolved proof obligations.

\section{The Grasshopper Problem as a Formalisation Target}

We first state the mathematical theorem in its standard form.

\begin{theorem}[Grasshopper problem]
Let \(n\) be a positive integer. Let \(a_1,\ldots,a_n\) be distinct positive integers, and let \(M\) be a set of \(n-1\) positive integers such that
\[
  a_1+\cdots+a_n \notin M.
\]
Then there exists a permutation \(\sigma\) of \(\{1,\ldots,n\}\) such that
\[
  a_{\sigma(1)}+\cdots+a_{\sigma(k)} \notin M
\]
for every \(1\le k\le n\).
\end{theorem}

\begin{remark}
Theorem~1 is stated to identify the mathematical target of the formalisation attempt. The present paper does not give a new proof of this theorem, and the accompanying Lean artefact does not contain a completed proof of it.
\end{remark}

The final condition at \(k=n\) is guaranteed by the hypothesis on the total sum, since the endpoint is independent of the ordering. The difficulty is to choose an ordering for which the first \(n-1\) partial sums avoid \(M\). The positivity of the jumps ensures that landing positions move strictly to the right, while distinctness prevents certain exchange operations from becoming degenerate.

The Lean statement generated in the reported artefact slightly generalises the original IMO formulation in two ways. First, its Lean cardinality hypothesis corresponds to \(|M|<n\), rather than \(|M| = n-1\). This is a more general theorem statement, since it allows any forbidden set of cardinality less than \(n\). Second, it uses \texttt{M : Finset Nat}, so zero may belong to \(M\). This does not affect the original problem, because all landing positions after at least one jump are positive under the hypothesis \(0 < a_i\).

Although the informal problem assumes \(n>0\), the Lean statement allows \(n=0\). In that case the cardinality hypothesis \(|M|<n\) is impossible, so the theorem is vacuous.

\section{Exchange Proof Strategy}

For a permutation \(\sigma\), define the partial sums
\[
  S_k(\sigma)=a_{\sigma(1)}+\cdots+a_{\sigma(k)},\qquad 1\le k\le n.
\]
Call an index \(k\) safe for \(\sigma\) if \(S_k(\sigma)\notin M\), and unsafe if \(S_k(\sigma)\in M\). The Lean development defines a score \(G(\sigma)\) as the sum of the non-forbidden partial sums and proves a local exchange lemma conditional on a permutation \(\sigma\) satisfying a maximality hypothesis for this score. Under this maximality hypothesis, if the permutation has an unsafe index, one swaps adjacent jumps whose order affects the corresponding intermediate partial sum and compares the resulting partial sums.

The local calculation is straightforward. Suppose two neighbouring jumps \(x\) and \(y\) occur after a previous partial sum \(P\). Before the swap the intermediate landing is \(P+x\), and after the swap it is \(P+y\). All other relevant partial sums are unchanged. If the original ordering was chosen to maximise \(G\), then a swap that increases the contribution at the changed partial sum while leaving the other contributions unchanged contradicts maximality. Thus maximality can force additional values of the form \(P+y\) to lie in \(M\).

The unresolved mathematical difficulty identified by the Aristotle output summary is the final counting argument. Completing the argument would require organising the values produced by many possible swaps and proving that they yield sufficiently many distinct elements of \(M\) to contradict \(|M|<n\). A naive induction also runs into a size problem: removing one jump need not reduce the forbidden set from size \(n-1\) to size \(n-2\). In the Lean source itself, however, the main theorem is closed directly by \texttt{sorry}; the file does not formally reduce the theorem to a separately encoded global counting lemma.

\section{AI-Assisted Lean Formalisation}

The formalisation attempt was conducted using the Aristotle API with Lean 4 and Mathlib~\cite{lean4,mathlib}. Aristotle is presented by Harmonic as a formal reasoning agent that can work from English mathematical input or inside an existing Lean project~\cite{aristotleweb}. The public SDK documentation describes an API and command-line workflow for interacting with Lean files~\cite{aristotlelib}.

Listing~\ref{lst:grasshopper} is a readability-oriented transcription of the theorem declaration. The accompanying Lean source remains authoritative for exact syntax, imports, notation, and proof status. In the listing, \texttt{notin} is used only as an ASCII rendering of Lean's non-membership notation \(\notin\).

\begin{lstlisting}[style=lean,caption={ASCII-rendered transcription of the Aristotle-reported Lean theorem declaration},label={lst:grasshopper}]
theorem grasshopper (n : Nat) (a : Fin n -> Nat) (M : Finset Nat)
    (ha_pos : forall i, 0 < a i)
    (ha_inj : Injective a)
    (hM_card : M.card < n)
    (hS : (sum i, a i) notin M) :
    exists sigma : Equiv.Perm (Fin n),
      forall k : Fin n, PS a sigma k notin M := by
  sorry
\end{lstlisting}

The use of \(\texttt{Fin n}\) represents jump indices as a finite type. A permutation of jumps is represented by \(\texttt{Equiv.Perm (Fin n)}\). The helper definition \(\texttt{PS}\) represents partial sums over initial segments such as \(\texttt{Finset.Iic k}\). This encoding is standard in Lean, but it requires explicit lemmas connecting final partial sums, adjacent swaps, and finite-set sums.

The finite type \(\texttt{Fin n}\) is zero-indexed. Thus the formal index \(\texttt{k : Fin n}\) indexes a landing position in the ordered sequence; it is not a literal rendering of the informal index \(1 \leq k \leq n\). When \(n > 0\), the first element of \(\texttt{Fin n}\) corresponds to the first landing position, and the last element corresponds to the final landing position.

\section{Verified Local Proof Components}

Table~\ref{tab:status} summarises the verification status reported for the Aristotle-generated Lean development.

\begin{table}[H]
\centering
\begin{tabular}{@{}>{\raggedright\arraybackslash}p{0.30\textwidth}>{\raggedright\arraybackslash}p{0.38\textwidth}>{\raggedright\arraybackslash}p{0.24\textwidth}@{}}
\toprule
Lean component & Mathematical role & Status \\
\midrule
\texttt{PS\_last} & Final partial sum equals total sum & Verified \\
\texttt{PS\_swap} & Adjacent transposition can affect only one relevant intermediate partial sum & Verified \\
\texttt{PS\_swap\_eq} & Computes the changed partial sum after an adjacent swap & Verified \\
\texttt{maximizer\_swap\_in\_M} & Converts maximality into a forbidden-set membership fact & Verified \\
\texttt{grasshopper} & Main theorem & Incomplete, one \texttt{sorry} \\
\bottomrule
\end{tabular}
\caption{Verification status of the Aristotle-generated Lean development.}
\label{tab:status}
\end{table}

The verified helper lemmas correspond to the local portion of the exchange proof. The lemma \texttt{PS\_last} establishes that the final formal partial sum agrees with the total sum. The lemmas \texttt{PS\_swap} and \texttt{PS\_swap\_eq} formalise the effect of an adjacent transposition: such a swap can affect only the relevant intermediate partial sum, and the changed partial sum has the expected value. The lemma \texttt{maximizer\_swap\_in\_M} formalises a maximality argument. If a chosen permutation maximises \(G\) and the partial sum at a position admitting an adjacent successor swap is forbidden, then the value obtained by that adjacent swap must also be forbidden. These lemmas verify the local exchange calculations needed to track partial sums under adjacent transpositions. They formalise the part of the exchange argument where the informal proof relies on the claim that adjacent swaps preserve all partial sums except the relevant intermediate one. They do not imply the Grasshopper theorem.

\section{The Intended Global Counting Argument}

The Aristotle output summary identifies the intended remaining mathematical step as the final counting or contradiction argument. Informally, the verified maximality lemma produces new membership facts in \(M\) at positions admitting adjacent successor swaps. To finish the theorem, one would have to show that these facts force at least \(n\) distinct forbidden values, contradicting \(|M|<n\).

This step is global. It must track possible collisions among values obtained from different swaps, use injectivity of the jump-length function, and relate the resulting set of values to the finite-cardinality assumption. In the Lean source itself, however, the main theorem \texttt{grasshopper} is closed directly by \texttt{sorry}. Thus, the artefact should be read as four verified helper lemmas together with an unproved main theorem, rather than as a formal Lean proof reduced to a separately encoded counting lemma.

\section{Reproducibility and Artefact Inspection}

The accompanying Lean archive contains the following project files and supporting documentation.
\begin{itemize}[leftmargin=*]
  \item \texttt{ARISTOTLE\_SUMMARY.md}
  \item \texttt{README.md}
  \item \texttt{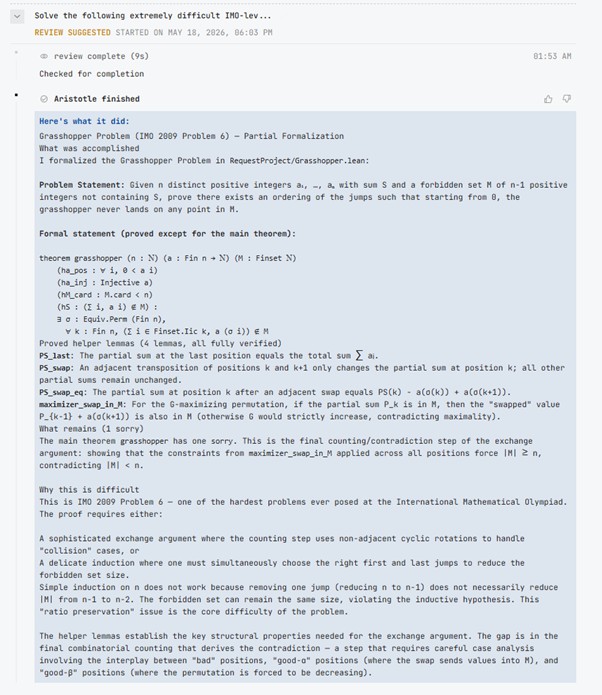}
  \item \texttt{RequestProject/Grasshopper.lean}
  \item \texttt{RequestProject/Main.lean}
  \item \texttt{lean-toolchain}
  \item \texttt{lakefile.toml}
  \item \texttt{lake-manifest.json}.
\end{itemize}

\begin{figure}[htbp]
\centering
\includegraphics[width=\textwidth,height=0.55\textheight,keepaspectratio]{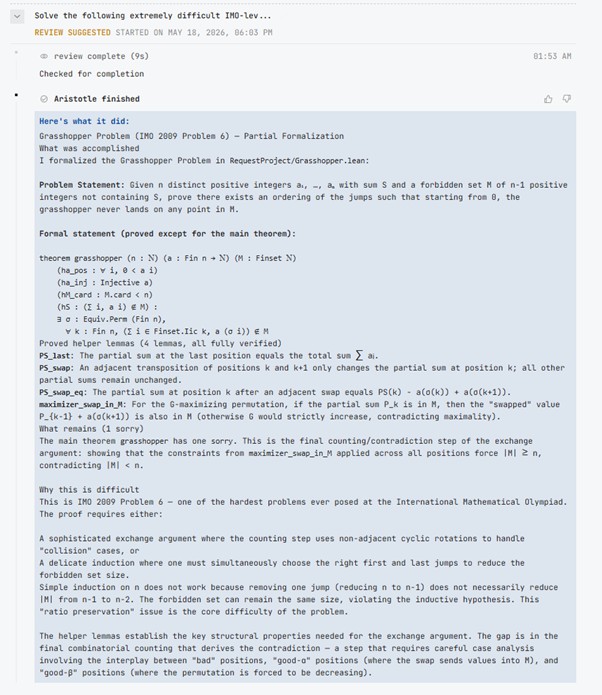}
\caption{Aristotle run log for the reported partial formalisation attempt.}
\label{fig:aristotle-run-log}
\end{figure}

The visible Aristotle run log indicates an approximate runtime of eight hours. Figure~\ref{fig:aristotle-run-log} shows the visible run-log evidence for this approximate runtime.

\section{Implications for AI-Assisted Theorem Proving}

The case study separates two kinds of formalisation work. Aristotle successfully handled local reasoning involving finite sums, permutations, adjacent swaps, and a maximality lemma. These tasks are amenable to proof-assistant verification once the relevant definitions are in place.

The generated development did not prove the main theorem. The Aristotle output summary identifies the intended missing mathematical step as the global counting or contradiction argument, which would require coordinating maximality, injectivity, positivity, and finite-cardinality estimates across many candidate partial sums. The Lean source itself does not encode this as a separate final subgoal; it closes \texttt{grasshopper} directly by \texttt{sorry}. Accordingly, the missing argument should be treated as an unproved obligation, not as a completed machine-checked argument.

For AI-assisted theorem proving, the lesson is methodological. A formal artefact should be inspected at the level of declarations, dependencies, placeholders, and proof obligations. This distinction matters for trustworthy AI because generated proof attempts may combine verified components with unresolved obligations. The artefact is useful as a case study because its verification status is inspectable. It contains verified local components and a clearly identified missing proof obligation.

\section*{Code availability}
The supplementary archive \texttt{grasshopper-lean-artifact.zip}, provided with this paper as supplementary material in the arXiv source package, contains the Lean project. To inspect the artefact, unzip the archive and run \texttt{lake build} from the project root. The main theorem \texttt{grasshopper} remains closed by \texttt{sorry}; the helper lemmas listed in Table~\ref{tab:status} were inspected separately.

\appendix

\section{Prompt Given to Aristotle API}

\begin{lstlisting}[style=shell]
Solve the following extremely difficult IMO-level combinatorics problem rigorously.

Let n be a positive integer. Let a_1, a_2, ..., a_n be distinct positive integers, and let M be a set of n - 1 positive integers such that

a_1 + a_2 + ... + a_n is not in M.

A grasshopper starts at position 0 on the number line. It will make n jumps to the right, using the jump lengths a_1, a_2, ..., a_n in some order, each exactly once.

Prove that there exists an ordering of the jumps such that the grasshopper never lands on any point in M.

Requirements:
1. Give a complete proof, not just a sketch.
2. Do not cite this as a known theorem or IMO problem.
3. Prove all auxiliary lemmas used.
4. The proof must work for all positive integers n.
5. The argument must explicitly construct or justify the existence of a valid permutation of the jump lengths.
6. Avoid computational checking except for trivial base cases.
7. If using induction, carefully justify the reduction step, especially how the forbidden set changes.
8. If using a maximality, exchange, or graph-theoretic argument, clearly define the object being optimised and prove why the chosen object must satisfy the required avoidance property.
\end{lstlisting}

\section{Aristotle Output Summary}

This appendix records the final Aristotle output summary for provenance. The mathematical interpretation of these claims is given in Sections 4 to 7.

\subsection*{Grasshopper Problem, partial formalisation}

Aristotle reported that it formalised the Grasshopper problem in \texttt{RequestProject/Grasshopper.lean}. The reported theorem declaration can be rendered schematically in ASCII as follows.

\begin{lstlisting}[style=lean]
theorem grasshopper (n : Nat) (a : Fin n -> Nat) (M : Finset Nat)
    (ha_pos : forall i, 0 < a i)
    (ha_inj : Injective a)
    (hM_card : M.card < n)
    (hS : (sum i, a i) notin M) :
    exists sigma : Equiv.Perm (Fin n),
      forall k : Fin n, PS a sigma k notin M := by
  sorry
\end{lstlisting}

Aristotle reported four helper lemmas as fully verified.
\begin{itemize}[leftmargin=*]
  \item \texttt{PS\_last}. The final partial sum \(S_n(\sigma)\) equals the total sum \(\sum a_i\).
  \item \texttt{PS\_swap}. An adjacent transposition of positions \(k\) and \(k+1\) can affect only the intermediate partial sum \(S_k(\sigma)\); all other partial sums \(S_j(\sigma)\) are unchanged.
  \item \texttt{PS\_swap\_eq}. The partial sum at position \(k\) after an adjacent swap equals \(S_k(\sigma)-a_{\sigma(k)}+a_{\sigma(k+1)}\).
  \item \texttt{maximizer\_swap\_in\_M}. For a permutation satisfying the \(G\)-maximality hypothesis, if the partial sum \(S_k(\sigma)\) at a position with an adjacent successor swap is in \(M\), then the swapped value \(S_k(\sigma)-a_{\sigma(k)}+a_{\sigma(k+1)}\) is also in \(M\).
\end{itemize}

The main theorem \texttt{grasshopper} is closed directly by one \texttt{sorry}. The Aristotle output summary identifies the intended remaining mathematical step as the final counting or contradiction argument, where one would show that the constraints from \texttt{maximizer\_swap\_in\_M} force \(|M|\ge n\), contradicting \(|M|<n\).

\bibliographystyle{plain}
\begingroup
\sloppy
\hbadness=10000
\bibliography{refs}
\endgroup

\end{document}